\documentclass[letterpaper]{article} %
\usepackage{aaai24}  %
\usepackage{times}  %
\usepackage{helvet}  %
\usepackage{courier}  %
\usepackage[hyphens]{url}  %
\usepackage{graphicx} %
\usepackage{natbib}  %
\usepackage{caption} %
\usepackage{algorithm}
\usepackage{algorithmic}

\usepackage{newfloat}
\usepackage{listings}
\DeclareCaptionStyle{ruled}{labelfont=normalfont,labelsep=colon,strut=off} %
\floatstyle{ruled}
\newfloat{listing}{tb}{lst}{}
\floatname{listing}{Listing}

\usepackage{booktabs}
\usepackage{multirow}
\usepackage{subcaption}
\usepackage{tabularx}
\usepackage{array}
\usepackage{xcolor}

\newcommand{\Method}{LAVE}

\title{Improving Automatic VQA Evaluation Using Large Language Models}
\author {
    Oscar Mañas\textsuperscript{\rm 1,2},
    Benno Krojer\textsuperscript{\rm 1,3},
    Aishwarya Agrawal\textsuperscript{\rm 1,2}
}
\affiliations {
    \textsuperscript{\rm 1}Mila
    \textsuperscript{\rm 2}Université de Montréal
    \textsuperscript{\rm 3}McGill University\\[6pt]
    \texttt{oscar.manas@mila.quebec}\\[10pt]
}

\begin{document}

\maketitle

\begin{abstract}
8 years after the visual question answering (VQA) task was proposed, accuracy remains the primary metric for automatic evaluation. VQA Accuracy has been effective so far in the IID evaluation setting. However, our community is undergoing a shift towards open-ended generative models and OOD evaluation. In this new paradigm, the existing VQA Accuracy metric is overly stringent and underestimates the performance of VQA systems. Thus, there is a need to develop more robust automatic VQA metrics that serve as a proxy for human judgment. In this work, we propose to leverage the in-context learning capabilities of instruction-tuned large language models (LLMs) to build a better VQA metric. We formulate VQA evaluation as an answer-rating task where the LLM is instructed to score the accuracy of a candidate answer given a set of reference answers. We demonstrate the proposed metric better correlates with human judgment compared to existing metrics across several VQA models and benchmarks. We hope wide adoption of our metric will contribute to better estimating the research progress on the VQA task. We plan to release the evaluation code and collected human judgments.\looseness-1
\end{abstract}

\section{Introduction}

\begin{figure}[t]
    \centering
    \includegraphics[width=0.47\textwidth]{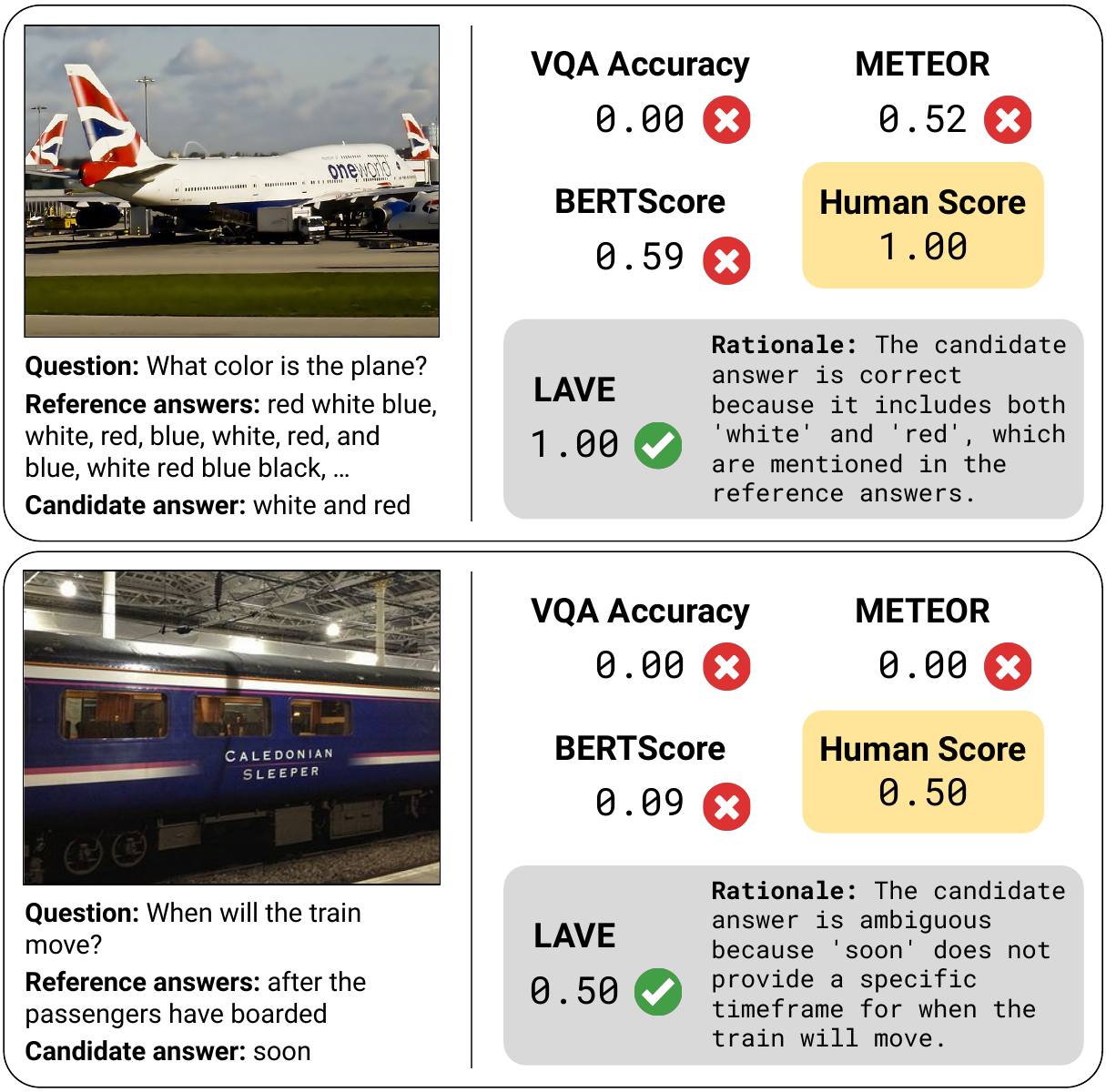}
    \caption{Existing VQA metrics and other strong baselines tend to miss out on correct answers generated by VQA models. Our proposed metric, \Method{}, is more aligned with human judgment and provides a rationale for its rating, making it also more interpretable.}
    \label{fig:teaser}
\end{figure}

Visual question answering (VQA)~\citep{antol2015vqa} has become an essential benchmark for assessing the progress of multimodal vision-language systems. 8 years after the task was proposed, accuracy remains the primary metric for automatically evaluating model performance. VQA Accuracy is based on exact string matching between a candidate answer predicted by the model and a set of reference answers annotated by humans. 
As pointed out in \citet{agrawal2023reassessing}, this metric has been effective so far because the VQA evaluation primarily followed the the independent and identically distributed (IID) paradigm, where the training and testing data distributions are quite similar. Thus, models could learn to adapt to the test answer distribution. However, recently, our community has been shifting its focus towards out-of-distribution (OOD) evaluation, either via zero-shot transfer to unseen VQA tasks or via finetuning on one VQA dataset and evaluating on another~\citep{agrawal2023reassessing}. In these settings, the answers generated by VQA models might not match any of the reference answers, while still being correct answers to the question! For instance, the generated answer might differ from the reference answers due to the format, specificity, different interpretations of the question, etc. (Sec.~\ref{sec:vqa_acc_failures}).
To address this limitation, some recent methods~\citep{li2023blip} have attempted to artificially modify the format of generated answers to align with the reference answers. However, we argue that this adjustment is a consequence of the flawed evaluation metric and should not influence modeling. Although human evaluation is the most reliable method for assessing generative models, it is costly and not scalable. Thus, there is a need to develop more robust automatic VQA metrics that serve as a proxy for human judgment.\looseness-1

A potential solution towards this issue would be to use soft evaluation metrics based on answer similarity (e.g., BERTScore~\citep{zhangbertscore}, Sentence-BERT~\citep{reimers2019sentence}). While these metrics might be effective in matching synonyms and paraphrases, they fail when the compared texts have fine-grained yet major differences (e.g., ``the man on the \textit{left}'' vs. ``the man on the \textit{right}'', ``there \textit{is a} dog'' vs. ``there \textit{is no} dog''). We empirically evaluated the performance of such soft metrics for VQA, and found that their correlation with human judgement is even weaker than that of VQA Accuracy (Sec.~\ref{sec:experiments}).

Inspired by recent advances in using large language models (LLMs) to evaluate natural language generation (NLG)~\citep{fu2023gptscore,liu2023gpteval,zheng2023judging}, we explore the potential of leveraging LLMs as superior evaluators of answer quality in VQA. We believe that LLMs have the potential to capture human preference given their extensive training in modeling human language, and hence present a compelling choice as proxy for human judgment. By employing LLMs, we can harness the benefits of soft metrics while mitigating their limitations, resulting in a more robust evaluation framework.

To this end, we propose a novel automatic VQA evaluation metric, \textit{\Method{}} (\textbf{L}LM-\textbf{A}ssisted \textbf{V}QA \textbf{E}valuation), which leverages the in-context learning capabilities of instruction-tuned LLMs. In particular, we formulate VQA evaluation as an answer-rating task where the LLM is instructed to score the correctness of a candidate answer given the corresponding question and a set of reference answers. To evaluate the effectiveness of the proposed metric, we collect human judgments on the correctness of answers generated by several state-of-the-art VQA models across three popular VQA benchmarks. Our results demonstrate that \Method{} correlates better with human judgment compared to existing metrics in diverse settings (Fig.~\ref{fig:teaser}). We also systematically categorize the failure modes of VQA Accuracy and show that \Method{} is able to recover most missed correct candidate answers. In addition, we conduct ablation studies to assess the impact of each design choice on the performance of \Method{}. In summary, our contributions are:

\begin{itemize}
    \item We propose a novel metric for automatic VQA evaluation, \Method{}, leveraging the in-context learning capabilities of instruction-tuned LLMs.
    \item We rigorously assess the effectiveness of \Method{} by computing its correlation with human judgment, and show its robustness across various VQA models and benchmarks.
    \item We benchmark several strong baseline metrics in addition to VQA Accuracy, such as BERTScore or S-BERTScore, and show \Method{} outperforms all of them.
    \item We systematically categorize the failure modes of VQA Accuracy and show \Method{} fixes most of its pitfalls.
    \item We conduct thorough ablation experiments to measure the effect of each component of \Method{}.
\end{itemize}

\section{Related Work}

\paragraph{Metrics for VQA}
VQA evaluation has received limited attention since the original VQA Accuracy metric was introduced by \citet{antol2015vqa}. In a later study, \citet{luo2021just} propose enhancing the reference answers with alternative answer sets (AAS), focusing on the case where only one reference answer per question is provided. More recently, \citet{hu2022promptcap} devise a soft VQA Accuracy metric as part of their data filtering pipeline. We implement this metric as one of our baselines for comparison (Sec.~\ref{sec:experimental_setup}).

\paragraph{Metrics for GenQA}
VQA and QA both involve answering questions related to a given context, either visual or textual. Generative QA evaluation faces similar challenges as VQA, relying primarily on metrics such as exact-match (EM) and F1 Score. Similarly to \citet{luo2021just}, \citet{si2021s} also propose expanding reference answers with equivalent ones mined from knowledge bases. \citet{lee2021kpqa} introduce a metric that weights answer tokens via keyphrase prediction. \citet{chen2019evaluating} found that a straightforward application of BERTScore fails to provide stronger correlation with human judgements. Instead, other works~\citep{risch2021semantic,bulian2022tomayto} train a semantic answer similarity metric based on BERT, showing improved correlation with human judgment. In contrast, we explore the capabilities of instruction-tuned LLMs in comparing candidate and reference answers.

\paragraph{Using LLMs as evaluators}
Recently, several works~\cite{fu2023gptscore,liu2023gpteval,kamalloo2023evaluating,li2023mimic,zheng2023judging,rajani2023llm_labels} have explored the possibility of using LLMs (Flan-T5~\citep{chung2022scaling}, OPT~\cite{zhang2022opt}, GPT-X~\cite{brown2020language,openai2023gpt}) to evaluate text generation for different tasks (e.g., summarization, dialogue generation, machine translation, QA, ...). Closer to our work, \citet{zhou2023lima} propose using ChatGPT to automatically evaluate model outputs on a Likert scale. However, the quality of their metric remains uncertain as they do not provide any evidence of its alignment with human judgment. In this work, we aim to rigorously assess the effectiveness of LLMs in evaluating VQA by measuring their correlation with human judgment in diverse settings.\looseness-1

\begin{table*}[t]
    \centering
    \caption{Failure modes of the strict VQA Accuracy where correct responses are marked as incorrect. Model generated answers are marked in \textcolor{blue}{blue} and reference answers in \textcolor{orange}{orange}.}
    \small
    \begin{tabular}{p{0.13\textwidth}p{0.27\textwidth}p{0.45\textwidth}p{0.05\textwidth}}
        \toprule
        \textbf{Category} & \textbf{Definition} & \textbf{Examples} & \textbf{\%} \\
        \midrule
        Multiple answers & Subjective, answers might focus on different aspects of the scene/activity. & \textbf{Q}: What are the sticks for? \textbf{A}: balance, pushing, skating, ... (Fig.~\ref{fig:skating}) & $34.25$ \\
        \midrule
        Over- or under-specifying and verbosity & The candidate answer contains more/less details than the references or is more/less verbose. & \textbf{Q}: Where is the hydrant? \textbf{A}: \textcolor{blue}{on the right}, \textcolor{orange}{right}; \textbf{Q}: What color are the flowers? \textbf{A}: \textcolor{blue}{pink}, \textcolor{orange}{pink and orange and red} & $27.75$ \\
        \midrule
        Synonym & Includes ``almost-synonym`` relation. & \textbf{Q}: What is the setting of this picture? \textbf{A}: \textcolor{blue}{field}, \textcolor{orange}{plains, grassland}; \textbf{Q}: What is the sign telling you to do or not do? \textbf{A}: \textcolor{blue}{no entry}, \textcolor{orange}{do not enter} & $21.0$ \\
        \midrule
        Broad/bad question or generic response & Question is near impossible to answer or highly subjective; model avoids answering by being overly generic. & \textbf{Q}: How many sheep are there? \textbf{A}: \textcolor{blue}{many}; \textbf{Q}: What is the current condition of these animals? (image simply shows a baby elephant) & $18.0$ \\
        \midrule
        Incorrect & Human judgment is incorrect. & - & $8.25$ \\
        \midrule
        Same stem & Reference and candidate share the same stem (plural vs. singular or gerund); different formatting or whitespace. & \textbf{Q}: What are the people doing? \textbf{A}: playing video \textcolor{blue}{games}/\textcolor{orange}{game}; \textbf{Q}: What shape is the building? \textbf{A}: \textcolor{blue}{rectangular}, \textcolor{orange}{rectangle}; \textbf{Q}: What colors are in the surfer's shirt? \textbf{A}: \textcolor{blue}{blue and white}, \textcolor{orange}{white and blue} & $5.75$ \\
        \midrule
        Hypernym & ``Subcategory-of'' relation. & \textbf{Q}: What are the people doing? \textbf{A}: \textcolor{blue}{playing wii}, \textcolor{orange}{playing video games}; \textbf{Q}: What is in the blender? \textbf{A}: \textcolor{blue}{vegetables}, \textcolor{orange}{carrots} & $5.0$ \\
        \midrule
        Unknown issue &  Model responds with ``unknown''. & - & $3.75$ \\
        \midrule
        Ambiguous object & Phrase could refer to multiple objects in the image. & \textbf{Q}: What kind of sign is this? \textbf{A}: \textcolor{blue}{billboard}, \textcolor{orange}{street sign} (image shows multiple signs/billboards) & $2.0$ \\
        \bottomrule
    \end{tabular}
    \label{tab:failure_modes}
\end{table*}

\begin{figure}[b]
    \centering
    \includegraphics[width=0.3\textwidth]{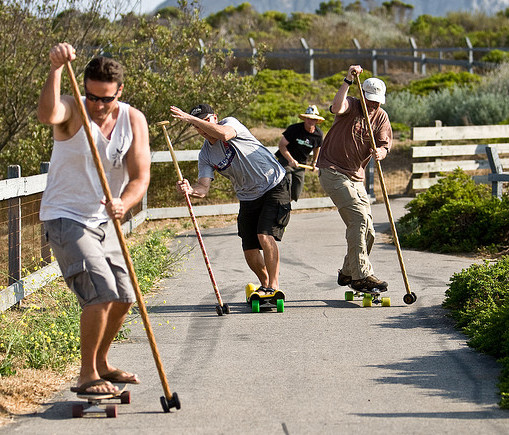}
    \caption{Example of a VQA Accuracy failure mode from the \textit{multiple answers} category (Tab.~\ref{tab:failure_modes}). Q: What are the sticks for? A: balance, pushing, skating, ...}
    \label{fig:skating}
\end{figure}

\section{Analysis of VQA Accuracy Failure Modes}
\label{sec:vqa_acc_failures}

The motivation for developing a new VQA metric arises from the limitations of VQA Accuracy in handling open-ended model responses, which are not suitable for exact string matching. To understand the specific failure modes a new metric should address, we conducted a small study where we manually categorized 400 VQA examples. We looked at examples where VQA Accuracy is below 0.5 (at most 1 out of 10 reference answers matches with the model's response), but at least 4 out of 5 humans rated the model's response as correct. In other words: when are actually correct responses marked as incorrect the way current VQA systems are evaluated? We annotated 100 examples for each of four model-dataset pairs: (BLIP-2, VQAv2), (BLIP-2, VG-QA), (BLIP-2, OK-VQA), and (PromptCap, VQAv2). We focus on BLIP-2 and PromptCap since their generation is most open-ended.\looseness-1

Our initial set of failure modes is inspired from \citet{luo2021just}, and manual inspection resulted in several additional categories. For clarity and conciseness, we decided to merge certain categories. Tab.~\ref{tab:failure_modes} shows the consolidated nine categories with definitions, examples and frequencies.

We identified four prevalent failure modes: (1) \textbf{multiple answers}, (2) \textbf{over- or under-specification and verbosity}, (3) \textbf{synonyms} and (4) \textbf{broad/bad question or generic response}. We observe that certain question types naturally lead to various possible correct answers. For instance, many where-questions (e.g., ``Where is the clock?'') can be answered using either absolute positioning or relative positioning to other objects. Other open-ended questions, such as asking what a person is doing or feeling, can be interpreted in multiple ways (e.g., ``Why is she posing for picture?''). \citet{luo2021just} introduced the category \textit{ambiguous object} when a phrase in the question could point to several objects (e.g., ``What color is the shirt?'' when there are several shirts). However, our inspection showed only a few occurrences of it and we speculate it often also falls into the \textit{multiple answers} category.\looseness-1

In summary, our analysis revealed that the open-ended nature of visual question answering can lead to multiple complex failure modes in VQA Accuracy.

\begin{figure*}[ht]
    \centering
    \includegraphics[width=0.99\textwidth]{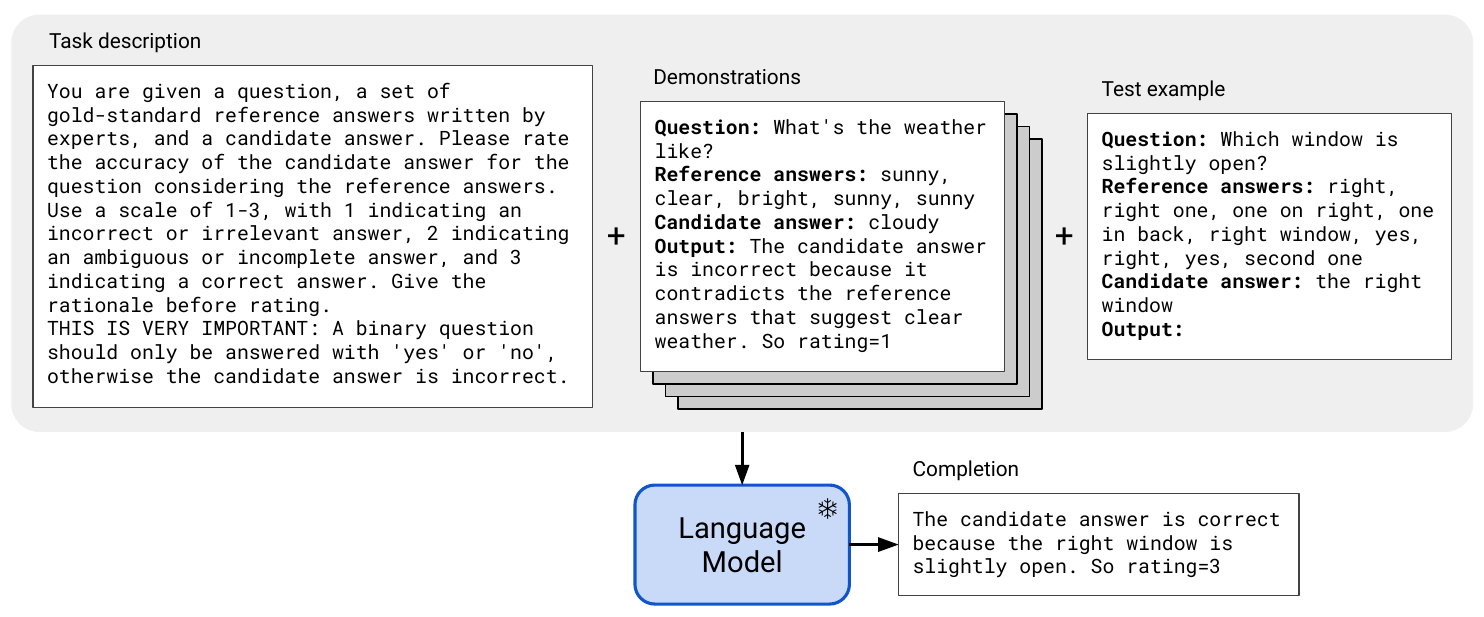}
    \caption{VQA evaluation with an LLM via in-context learning.}
    \label{fig:prompt}
\end{figure*}

\section{Method}

We present \Method{}, an LLM-based evaluation framework to automatically assess the quality of answers generated by VQA models. Each VQA example comprises an image $i$, a question about the image $q$, and a set of reference answers $R$ provided by human annotators. Given $i$ and $q$, a VQA model $f$ generates a candidate answer $c$, i.e., $c = f(i, q)$. Our goal is to automatically evaluate the quality of the generated answer $c$ by comparing it with the references $R$. To enhance the evaluation process, we can additionally leverage the contextual information from the question $q$ and the image $i$. We build a textual prompt using $R$, $c$, and optionally $q$ and $i$ (as an image description). This prompt is then fed to an LLM to generate a quality rating. The following sections describe the key design decisions underlying our approach.

\subsection{Choosing a Large Language Model}

It is crucial to choose an appropriate LLM as \Method{}'s performance directly hinges on its capabilities. We pose VQA evaluation as a close-ended answer-verification task and adapt a frozen LLM through in-context learning. Hence, we opt for an instruction-tuned LLM, which has demonstrated superior performance in transferring to new tasks with limited demonstrations~\citep{weifinetuned}. Instruction-tuned LLMs are also more robust to prompt selection, and they can match the few-shot performance of much larger LLMs pretrained with self-supervised objectives.
Considering all these factors, we first select the Flan-T5~\citep{chung2022scaling} model family for our metric. In addition, Flan-T5 is finetuned on chain-of-thought (CoT) data, enabling it to provide reasoning for its answers.

To demonstrate \Method{}'s robustness across different LLMs, we also consider Vicuna-v1.3~\citep{chiang2023vicuna} and GPT-3.5-Turbo (aka ChatGPT~\citep{openai2022chatgpt}). We optimize our prompt for Flan-T5 (Sec.~\ref{sec:ablations}) and subsequently use the same prompt with the other LLMs. This opens the possibility of enhancing our metric without extra effort as stronger LLMs become available in the future.

\subsection{Prompt for VQA Evaluation}

We frame VQA evaluation as an answer-rating task amenable to in-context learning with LLMs. We adopt a rating scale ranging from 1 to 3 (as opposed to a binary rating) to account for ambiguous questions or incomplete answers. Our prompt (Fig.~\ref{fig:prompt}) comprises the typical components: task description, a few demonstrations of input/output, and the input for a test example. We draw inspiration from SNI~\cite{wang2022super} to structure our task description, as it is one of the main sources of training data for instruction-tuned LLMs. We also append ``\texttt{Give the rationale before rating.}'' to elicit a justification for the assigned rating, which improves explainability. Each demonstration consists of a question $q$, a set of reference answers $R$, the candidate answer $c$, the answer rating $r$, and an explanation $e$ for the rating. We observed binary questions are particularly challenging to evaluate (App.~\ref{sec:binary_questions}), so we manually curate two sets of 8 demonstrations, one for binary questions and the other for general questions. We ensure demonstrations are diverse and cover various question types, numbers of reference answers, levels of agreement, candidate answer precision and verbosity, ratings, etc (refer to App.~\ref{sec:demos} for the complete list). While these sets of demonstrations are designed to be comprehensive, users of our metric could also provide their own demonstrations to cover different cases. Additionally, to account for noise in the annotations, we filter out outlier reference answers that have a frequency lower than 25\% of the maximum answer frequency. Finally, the test example only includes $q$, $R$ and $c$, and the LLM is expected to generate an explanation $e$ followed by a rating $r$. We found incorporating visual context from the image $i$ into the prompt does not provide significant benefits (see Sec.~\ref{sec:ablations} and App.~\ref{sec:visual_context} for more details).

\subsection{Scoring Function}

Given the LLM's generated text for the test example, we extract the rating $r$ from the last character (either 1, 2 or 3) and linearly map it to a score $s$ in the range $[0, 1]$: $s = (r - 1) / 2$.
Inspired by \citet{liu2023gpteval}, we also explored the possibility of using the probabilities of output tokens to normalize the ratings and take their weighted sum as the final rating, but we did not observe any improvements in our task.

\section{Experiments}
\label{sec:experiments}

\subsection{Experimental Setup}
\label{sec:experimental_setup}

\paragraph{VQA models and benchmarks}
We evaluate \Method{} on answers generated by several VQA models across multiple VQA benchmarks. In particular, we consider two representative state-of-the-art VQA models: \textbf{BLIP-2 Flan-T5-XXL}~\citep{li2023blip} and \textbf{PromptCap GPT-3}~\citep{hu2022promptcap}. Our selection criteria were based on their public availability, their zero-shot VQA capability, and their architectural diversity. We also include \textbf{BLIP}~\citep{li2022blip} finetuned on VQAv2 (BLIP\textsubscript{VQA}) and VG-QA (BLIP\textsubscript{VG}), which represents the finetuning-OOD paradigm. We use these VQA models to generate answers for three VQA datasets: \textbf{VQAv2}~\cite{goyal2017making}, \textbf{VG-QA}~\citep{krishna2017visual} and \textbf{OK-VQA}~\cite{marino2019ok}. The selection of these datasets was driven by their diverse answer distributions. VQAv2 was prioritized due to its popularity as one of the most widely-used VQA benchmarks, providing 10 reference answers per question. VG-QA was chosen for its notably distinct answer distribution compared to VQAv2 (as shown by \citet{agrawal2023reassessing}), and it provides a single reference answer per question. Lastly, OK-VQA was selected for its unique answer distribution, differing from both VQAv2 and VG-QA, as some of its questions require external knowledge to answer.

\paragraph{Baselines}
We evaluate \Method{} against several strong baselines for VQA evaluation which involve comparing a candidate answer with a set of references. We consider the original \textbf{VQA Accuracy}~\cite{antol2015vqa}, based on exact string matching; \textbf{soft VQA Accuracy}~\cite{hu2022promptcap}, which replaces exact-match by edit distance (CER); \textbf{METEOR}~\cite{banerjee2005meteor}, which uses unigram matching on surface form, stemmed form and meaning; \textbf{CIDEr}~\cite{vedantam2015cider}, which captures consensus among multiple references; \textbf{BERTScore}~\cite{zhangbertscore}, which calculates pairwise cosine similarity of contextualized token embeddings; and \textbf{S-BERTScore}~\cite{reimers2019sentence}, which measures cosine similarity of sentence embeddings. Both BERTScore and S-BERTScore compute similarity between pairs of sentences, so when there are multiple reference answers, the maximum score with the candidate is selected.\looseness-1

\paragraph{Implementation details}
We consider Flan-T5-XXL and Vicuna-v1.3-13B as open-source LLMs, and GPT-3.5-Turbo (\texttt{gpt-3.5-turbo-0613}) as a closed-source LLM. We leverage the HuggingFace Transformers'~\cite{wolf2020transformers} implementation of Flan-T5 and LLaMA (for Vicuna), and use GPT-3.5-Turbo through OpenAI's API\footnote{\url{https://platform.openai.com/docs/api-reference}}. To make generation deterministic, we perform greedy decoding, or equivalently set the temperature to $0$ in OpenAI's API.

\subsection{Collecting Human Judgments}

We collected human judgments about answer quality using Amazon Mechanical Turk (MTurk) with the same web interface as \citet{agrawal2023reassessing}.
Specifically, we collected judgments for answers generated by BLIP-2 on VQAv2, VG-QA and OK-VQA,  PromptCap on VQAv2 and OK-VQA, BLIP\textsubscript{VQA} on VG-QA and OK-VQA, and BLIP\textsubscript{VG} on VQAv2 and OK-VQA ($2450$ questions each\footnote{We initially collected human judgments on $2500$ questions per model-dataset pair, but had to remove some to control data quality.\looseness-1}). In total, our test set contains $22.1$k questions.
We additionally collected validation/development sets of human judgments for answers generated by BLIP-2 on VQAv2 and VG-QA, and BLIP on VQAv2 and VG-QA ($1000$ questions each). In total, our validation set contains $4$k questions, which serve to guide our design choices. We emphasize that \textit{PromptCap and OK-VQA are completely unseen during metric development} to show \Method{}'s generality. Following \citep{agrawal2023reassessing}, each answer was assessed by 5 annotators who were asked to provide a binary rating (correct/incorrect) based on the corresponding image and question. An important difference between the task posed to turkers and to the LLM is that the LLM is provided with a list of reference answers annotated by humans, while turkers are not. 
After filtering out low-quality annotations, we obtain an inter-annotator agreement measured by Krippendorff's $\alpha$ of $62.0$. Upon manual inspection, we observed that model responses (and even questions) are often ambiguous in nature, which would explain the relatively low inter-annotator agreement (see App.~\ref{sec:crowdsourcing} for more details).
We derive a single ``quality'' score from the $5$ binary ratings per answer as follows: $1.0$ if at least $4$ annotators rate the answer as correct, $0.5$ if only $2$ or $3$ did so, and $0.0$ otherwise. Considering partial scores is crucial to acknowledge the ambiguity inherent in certain questions, particularly when dealing with generative models which can produce technically accurate answers that may not align with the intended meaning of the question.

\begin{table*}[ht]
    \centering
    \caption{Spearman correlation ($\rho$) between VQA metrics and human judgment.}
    \fontsize{9}{9.5}\selectfont
    \resizebox{0.99\textwidth}{!}{
    \begin{tabular}{lcccccccccc}
        \toprule
         & \multicolumn{3}{c}{BLIP-2} & \multicolumn{2}{c}{PromptCap} & \multicolumn{2}{c}{BLIP\textsubscript{VG}} & \multicolumn{2}{c}{BLIP\textsubscript{VQA}} & \multirow{2}{*}{Overall} \\
         & VQAv2 & VG-QA & OK-VQA & VQAv2 & OK-VQA & VQAv2 & OK-VQA & VG-QA & OK-VQA & \\
        \midrule
        \textbf{Baselines} \\
        VQA Acc. & 71.54 & 41.19 & 48.65 & 65.82 & 48.24 & 84.85 & 70.38 & 41.90 & 68.84 & 60.13 \\
        Soft VQA Acc. & 73.23 & 49.88 & 47.65 & 67.06 & 53.46 & 83.59 & 66.79 & 52.04 & 67.87 & 63.91 \\
        METEOR & 64.75 & 48.70 & 50.97 & 57.74 & 51.76 & 83.45 & \textbf{71.42} & 51.65 & 68.44 & 58.68 \\
        CIDEr & 69.55 & 47.78 & 53.23 & 63.26 & 49.81 & \textbf{85.07} & 71.08 & 46.50 & \textbf{70.25} & 63.88 \\
        BERTScore & 50.61 & 11.73 & 38.62 & 41.14 & 42.73 & 72.14 & 59.42 & 15.88 & 60.51 & 31.47 \\
        S-BERTScore & 60.44 & 42.10 & 47.84 & 47.44 & 47.11 & 77.65 & 68.04 & 47.61 & 65.61 & 56.61 \\
        \midrule
        \textbf{Ours} \\
        \Method{}~\textsubscript{FT5} & 71.19 & 59.94 & 59.85 & 64.18 & \textbf{58.67} & 71.67 & 66.03 & 54.50 & 63.87 & 64.99 \\
        \Method{}~\textsubscript{Vicuna} & 72.35 & 51.65 & 58.45 & 67.23 & 54.81 & 77.77 & \textbf{71.45} & 48.19 & 68.44 & 64.05 \\
        \Method{}~\textsubscript{GPT-3.5} & \textbf{74.25} & \textbf{60.19} & \textbf{61.47} & \textbf{71.99} & 57.39 & 83.63 & 69.97 & \textbf{58.47} & 67.57 & \textbf{68.91} \\
        \bottomrule
    \end{tabular}
    }
    \label{tab:spearman_results}
\end{table*}

\subsection{Correlation with Human Judgment}
\label{sec:correlation}

To evaluate \Method{}, we measure its correlation with human judgment using two widely accepted rank correlation coefficients: Spearman's $\rho$ and Kendall's $\tau$ (in the appendix). These provide a robust measure of the association between metrics, without assuming a linear relationship or a specific distribution of the data.
Both coefficients range from $-1$ to $1$, with $-1$/$+1$ meaning perfect inverse/direct correlation.

Spearman correlations between VQA metrics (ours and baselines) and human judgment on every evaluated pair of VQA model and dataset are shown in Tab.~\ref{tab:spearman_results} (see App.~\ref{sec:additional_results} for additional results and observations).
We verify statistical significance by bootstrapping with 5000 resamples and running a t-test with a significance level of 5\% between pairs of correlations. For each setting, we bold the best results which are significantly better than the second-best result.
The main observations are summarized as follows:

\textbf{Overall, \Method{} is significantly more aligned with human judgment than all the considered baselines}, independently of the underlying LLM. Among the considered LLMs, we observe GPT-3.5 provides the highest overall correlation with human judgment. This is in line with recent LLM leaderboards~\citep{zheng2023judging} which, at the time of writing, place GPT-3.5 (along GPT-4~\citep{openai2023gpt} and Claude~\citep{anthropic2023claude}) one step above any other LLM across a diverse set of benchmarks. However, we are excited to report that, on the VQA evaluation task, open-source LLMs such as Flan-T5 or Vicuna also outperform all baselines on average. Interestingly, despite being trained on user-shared conversations with ChatGPT, Vicuna falls behind Flan-T5 in our task.\looseness-1

\textbf{\Method{} generalizes to new VQA models and benchmarks.} We did not use human judgments of answers produced by PromptCap (across all datasets) or to OK-VQA questions (from all models) to guide our prompt design (see Sec.~\ref{sec:ablations}). Still, our metric correlates better with human judgment than all baselines in these hold-out settings. This indicates our design choices are not overfitted to the particular settings used during metric development, and that \Method{} appears promising for evaluating answers generated by various models and across different datasets.

\textbf{Questions from VQAv2 and OK-VQA answered by BLIP follow a different trend}. In these settings, VQA Accuracy's correlation with human judgment is considerably higher than for zero-shot VQA models, whereas \Method{} has only a slightly higher correlation.
We observe human score is much higher for BLIP-2 and PromptCap answers ($0.7552$ on average) compared to BLIP answers ($0.5293$ on average). 
Therefore, BLIP answers are more frequently incorrect or incomplete, which is expected as open-ended generative models are known to perform better on OOD data. The higher correlation for VQA Accuracy in these settings can be attributed to its efficacy in identifying incorrect candidate answers, while LLMs might label some as correct. For instance, GPT-3.5 labels ``sink`` as a correct answer to the question ``What is this kind of sink called?'', or ``refrigerator'' as a correct answer to ``What does this device generally do?''.
Thus, the different trend in correlation with human judgment can be explained by a higher frequency of incorrect answers. This trend does not hold for VG-QA because \Method{} outperforms the baselines when there is a single reference answer (see App.~\ref{sec:additional_results}).\looseness-1

\begin{table}[t]
    \centering
    \caption{Spearman correlation ($\rho$) between \Method{}~\textsubscript{FT5} and human judgment when ablating for prompt design choices.}
    \resizebox{0.47\textwidth}{!}{
    \begin{tabular}{lccccccc}
        \toprule
         & \multicolumn{2}{c}{BLIP-2} & \multicolumn{1}{c}{BLIP\textsubscript{VG}} & \multicolumn{1}{c}{BLIP\textsubscript{VQA}} & \multirow{2}{*}{Overall} \\
         & VQAv2 & VG-QA & VQAv2 & VG-QA & \\
        \midrule
        \Method{}~\textsubscript{FT5} & 67.50 & 61.57 & 74.82 & 63.09 & 66.74 \\
        \midrule
        1-shot & 55.43 & 61.04 & 59.79 & 57.45 & 59.07 \\
        4-shot & 68.92 & 60.30 & 73.37 & 60.50 & 65.40 \\
        \textit{w/o} rationale & 58.67 & 63.87 & 68.01 & 65.36 & 65.57 \\
        \textit{w/o} filter refs. & 62.53 & 61.58 & 74.09 & 63.05 & 64.89 \\
        \textit{w/} caption & 68.47 & 63.50 & 71.33 & 64.25 & \textbf{66.94} \\
        \bottomrule
    \end{tabular}
    }
    \label{tab:ablations}
\end{table}

\begin{table*}[t]
    \centering
    \caption{Selected questions from VQAv2 answered by BLIP-2, evaluated by VQA Accuracy and \Method{}~\textsubscript{GPT-3.5}, along with the rationale for the answer rating. Duplicate reference answers have been omitted for conciseness.}
    \small
    \begin{tabular}{lp{2.25cm}p{2.25cm}p{2.5cm}p{1.5cm}p{1.25cm}p{1.25cm}p{3cm}}
        \toprule
         & \multirow{2}{*}{Image} & \multirow{2}{*}{Question} & Reference & Candidate & VQA & \multirow{2}{*}{\Method{}~\textsubscript{GPT-3.5}} & \multirow{2}{*}{Rationale} \\
         & & & answers & answer & Accuracy & & \\
        \midrule
        \multirow{6}{*}{(a)} & \multirow{2}{*}{\includegraphics[height=2cm]{figures/COCO_val2014_000000383513.jpg}} & What kind of ecological system is this? & water, lake, unanswerable, tropical, lush, cloudy, warm & aquatic & 0.0 & 1.0 & The candidate answer is correct because it is synonymous with `water' and `lake', which are mentioned in the reference answers. \\
        \midrule
        \multirow{6}{*}{(b)} & \multirow{2}{*}{\includegraphics[height=2cm]{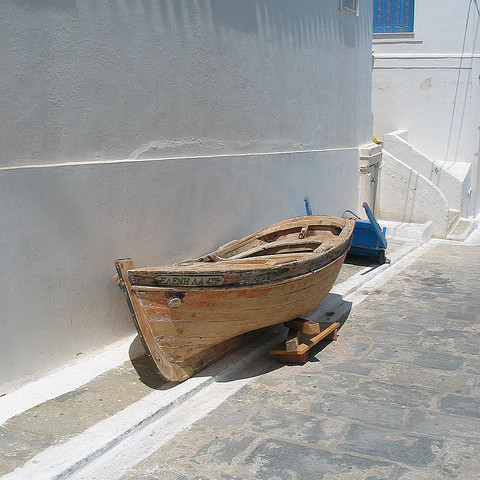}} & What is ironic about this photo? & no water, boat, boat is out of water, boat not in water, boat is not in water, boat on dry land, boat is on dry land, boat on land, boat is on land & there is a boat on the street & 0.0 & 1.0 & The candidate answer is correct because it captures the irony of the situation where a boat is on the street instead of in the water. \\
        \midrule
        \multirow{5}{*}{(c)} & \multirow{2}{*}{\includegraphics[width=2cm]{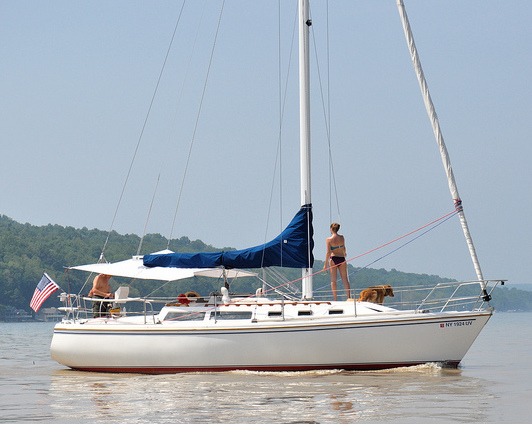}} & What country's flag is shown? & usa, united states of america & american & 0.0 & 1.0 & The candidate answer is correct because 'american' is equivalent to 'usa' in this context. \\
        \addlinespace\addlinespace\bottomrule
    \end{tabular}
    \label{tab:qualitative_analysis}
\end{table*}

\subsection{Ablation Studies}
\label{sec:ablations}

We compute correlation between \Method{}~\textsubscript{FT5} and human judgment when ablating for different prompt design choices. The best overall configuration is used to compute correlation on the test sets. Tab.~\ref{tab:ablations} summarizes our ablation results.

\paragraph{Number of demonstrations}
Our results suggest a positive correlation between the number of demonstrations and \Method{}~\textsubscript{FT5}'s effectiveness. As the number of demonstrations increases, the correlation with human judgment tends to improve. However, there is a tradeoff between number of demonstrations and computational overhead (and financial cost for GPT-3.5), so we tested up to 8 demonstrations.

\paragraph{Rationalization}
We measure the effect of asking the LLM to generate a rationale before rating candidate answers. Two trends arise when including rationalization: significantly improved performance on VQAv2 and slightly worse performance on VG-QA. We hypothesize a single reference answer (VG-QA) simplifies the answer-rating task, while having multiple reference answers (VQAv2) opens the door to discrepancies among annotators, leading to a more complex evaluation which can benefit from step-by-step reasoning.

\paragraph{Filtering of reference answers}
When using 10 reference answers (VQAv2), we observe that filtering out low-frequency answers consistently improves correlation, likely attributed to the reduction of noise in the reference answers. Note that this filtering has no effect when using a single reference answer (VG-QA).\looseness-1

\paragraph{Visual context}
We evaluate the effect of incorporating visual context into the prompt via an image description. Concretely, we use ground-truth captions from COCO and VG, modify the beginning of our task description as ``\texttt{You are given an image description, a question about the image, ...}'' and add ``\texttt{Image description: \{caption\}}'' to each example. Including visual context appears to be beneficial only in certain cases, especially when the dataset has a single reference answer (VG-QA). Notably, the overall correlation remains comparable to that observed without leveraging visual context. Moreover, in order to deploy our metric with visual context, we would need to add a captioning module to obtain the image description. Considering that visual context substantially increases the computational overhead due to image captioning and increased prompt length, we opted for excluding it from our final method.

\begin{figure}[t]
    \centering
    \includegraphics[width=0.47\textwidth]{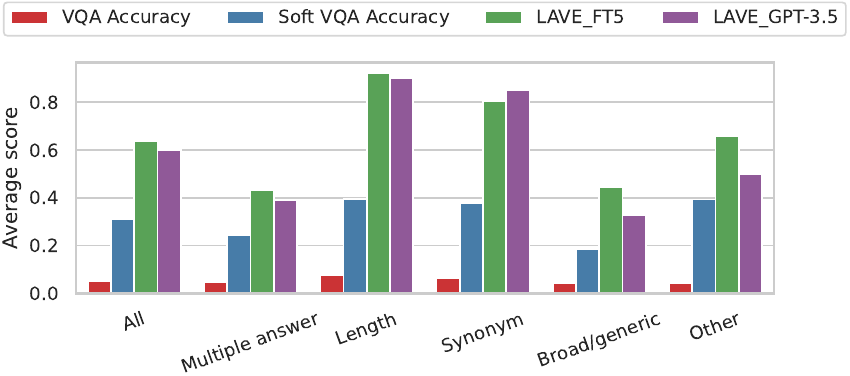}
    \caption{Average score of VQA evaluation metrics for cases where VQA Accuracy misses correct candidate answers, broken down by failure mode category.}
    \label{fig:vqa_acc_failures_analysis}
\end{figure}

\subsection{Does \Method{} Fix VQA Accuracy's Failures?}

Aside from having better overall correlation with human judgment, we would like to know how \Method{} behaves in the failure modes of VQA Accuracy highlighted in Sec.~\ref{sec:vqa_acc_failures}. As a reminder, these are all cases where human annotators collectively labeled the candidate answer as correct (score of $1.0$), while VQA Accuracy was below $0.5$ (either $0.3$ or $0.0$). Therefore, we would expect our metric to give these candidate answers a score of $1.0$ (excluding incorrect cases -- $8.25\%$)\footnote{Note that, in this setting, it is not possible to compute correlation with human judgment since it is constant ($1.0$).}. 
Out of the $22.1$k questions from our test sets, this is the case for $3601$ questions ($16.33\%$). For completeness, we found the reverse scenario, collective human score of $0.0$ and VQA Accuracy above $0.5$, occurs in 379 questions ($1.72\%$);
these are cases where the new human annotators disagree with the original annotations of the VQA datasets, indicating some noise in our collected human judgments.

Fig.~\ref{fig:vqa_acc_failures_analysis} shows the average score of VQA Accuracy, Soft VQA Accuracy, \Method{}~\textsubscript{FT5} and \Method{}~\textsubscript{GPT-3.5} on the 400 manually-labeled examples analyzed in Sec.~\ref{sec:vqa_acc_failures}. We observe that \Method{} is significantly more aligned with human judgment than both VQA Accuracy and Soft VQA Accurarcy, especially when candidate answers are more verbose or they are a synonym of the reference answers. As previously mentioned, broad questions or which have multiple correct answers may be overly subjective, so it is harder for an LLM to determine whether the candidate answer is correct. It is interesting to see, however, that in these cases Flan-T5 generally performs better than GPT-3.5. In summary, this indicates that \Method{} is able to recover a considerable fraction of correct candidate answers wrongly labeled as incorrect by VQA Accuracy.\looseness-1

Tab.~\ref{tab:qualitative_analysis} contains a few selected examples where \Method{}~\textsubscript{GPT-3.5} fixes failures of VQA Accuracy. For instance, example (a) shows our metric is able to identify that the candidate answer is a synonym of several references, even though the form is different. Example (b) demonstrates our metric is robust to answers of diverse verbosity. In example (c), our metric is capable of identifying the candidate answer is equivalent to the references, even though they belong to different lexical categories.

\section{Conclusions}

We present \Method{}, a new automatic VQA evaluation metric leveraging the in-context learning capabilities of instruction-tuned LLMs. Through a comprehensive study involving diverse VQA models and benchmarks, we demonstrate that \Method{} is significantly more aligned with human judgment compared to existing metrics. We hope wide adoption of our metric will contribute to better estimating the progress of vision-language systems on the VQA task.

\newpage

\section*{Ethical Statement}

In this work, we propose a novel VQA evaluation metric leveraging the power of instruction-tuned LLMs. While this advancement has the potential to significantly improve the evaluation and development of VQA systems, it also raises several ethical and societal considerations that warrant careful attention.\looseness-1

First, while \Method{} shows improved correlation with human judgment, we must acknowledge that the diversity and representativeness of the human annotators could influence the results. If the pool of annotators is not diverse, there may be biases in their judgments that could influence the performance of the proposed metric. We made a concerted effort to ensure that our pool of human annotators was as diverse as possible, but further research and mitigation strategies may be necessary to address this concern fully.

Second, the use of LLMs in any context brings up the issue of potential biases encoded in these models. As LLMs are typically trained on large-scale datasets scraped from the internet, they can inadvertently learn and perpetuate harmful biases present in those datasets. Such biases could result in discriminatory or otherwise unethical outcomes, so it is crucial to consider them when deploying or further developing \Method{}. Future work should continue to investigate methods to identify and mitigate these biases.

Lastly, it is important to consider the broader impact of our research on society, particularly as it relates to the automation of tasks traditionally performed by humans. While improving VQA evaluation metrics could lead to more efficient and accurate systems, the potential displacement of jobs traditionally performed by humans could have significant societal impacts. It is essential to consider these potential consequences and to work towards solutions that leverage the benefits of AI while also considering the human factor.\looseness-1

\section*{Acknowledgments}

We are grateful to Mila's IDT team for their technical support with the computational infrastructure. The authors acknowledge the material support of NVIDIA in the form of computational resources. During this project, Aishwarya Agrawal was supported by the Canada CIFAR AI Chair award. We would also like to thank Samsung Electronics Co., Ldt. for funding this research.

\bibliography{references}

\newpage

\appendix
\section{Appendix}

\subsection{Human Judgment Crowdsourcing}
\label{sec:crowdsourcing}

\begin{figure}[ht]
    \begin{subfigure}[b]{0.47\textwidth}
        \includegraphics[width=\textwidth]{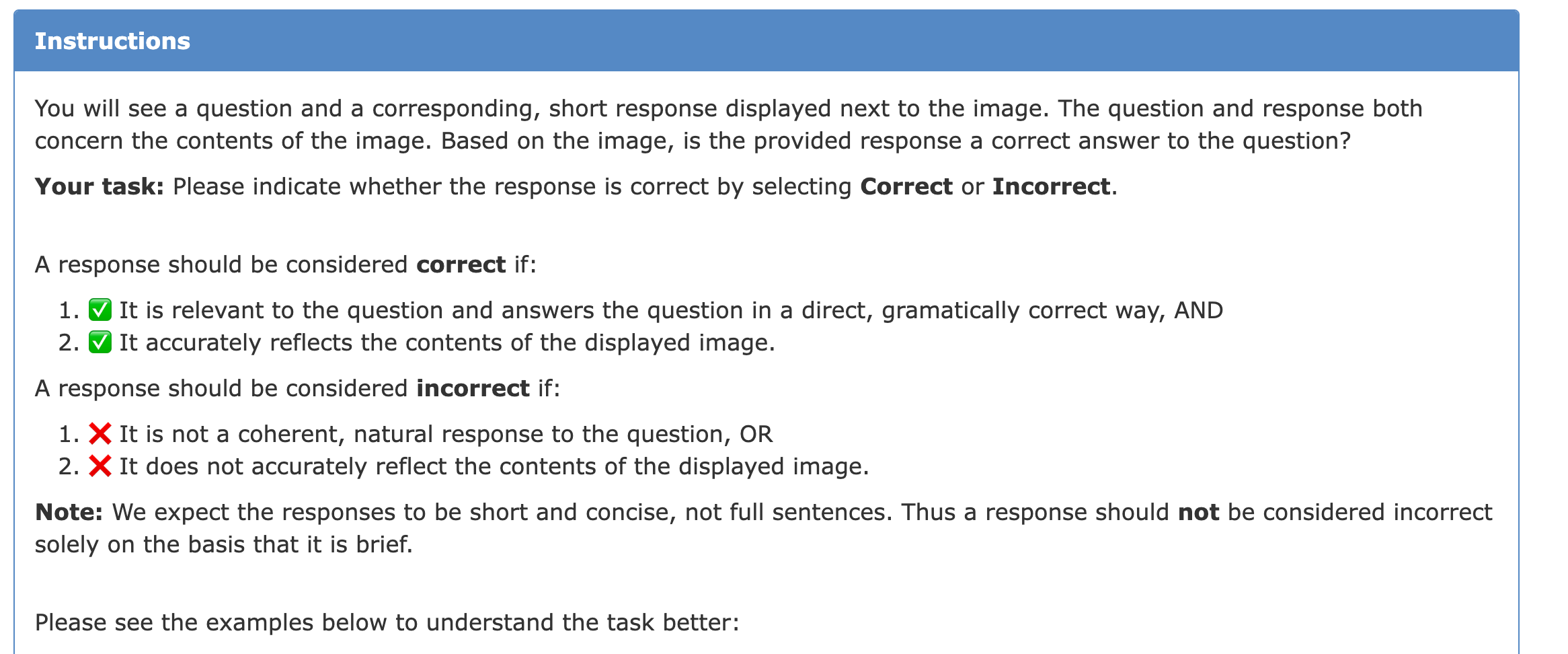}
        \caption{Main task instructions}
        \label{fig:subfig1}
    \end{subfigure}
    \hfill
    \begin{subfigure}[b]{0.47\textwidth}
        \includegraphics[width=\textwidth]{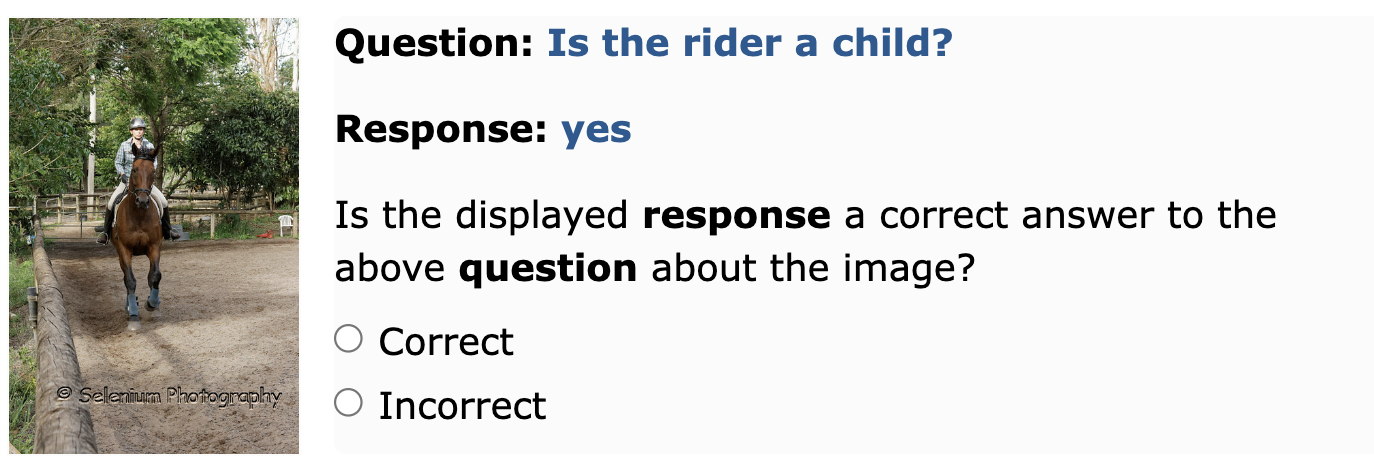}
        \caption{Example of a task}
        \label{fig:subfig2}
    \end{subfigure}
    \caption{Amazon Mechanical Turk interface}
    \label{fig:mturk_interface}
\end{figure}

An example of our annotator interface is shown in Fig.~\ref{fig:mturk_interface}.

We required MTurk workers to meet the following criteria: (1) a HIT Approval rate exceeding $0.98$, (2) over $1000$ HITs approved, and (3) location in the US or Canada. For their efforts, workers were compensated at a rate of \$$0.15$ per batch of 10 questions (\$$0.2$ for OK-VQA as certain questions need fact-checking).

\paragraph{Annotator Agreement.} After assessing the quality of workers and manually analyzing their output, we attribute the low Krippendorff's $\alpha$ value of $62.0$ to issues with inherent noise in the dataset and ambiguities in both questions and answers. To support this claim, we randomly sampled and visualized 10 examples of maximal disagreement, i.e. where 2 or 3 out of 5 annotators marked the model's response as correct (Tab.~\ref{tab:annotator_disagreement}).
While in some instances annotators were too lenient at interpreting the model's response, other times we believe it is genuinely difficult to decide if the response deserves partial credit.

\begin{table*}[ht]
    \centering
    \caption{Examples of maximal annotator disagreement.}
    \begin{tabular}{p{0.15\textwidth}p{0.28\textwidth}p{0.08\textwidth}p{0.08\textwidth}p{0.2\textwidth}}
        \toprule
        Image & Question & Candidate answer & Annotator rating & Comment \\
        \midrule
        \multirow{2}{*}{\includegraphics[height=2cm]{figures/COCO_val2014_000000463865.jpg}} & What breed it this cat? & kitten & 2/5 & - \\\\\\\\\\
        \addlinespace\midrule
        \multirow{2}{*}{\includegraphics[height=2cm]{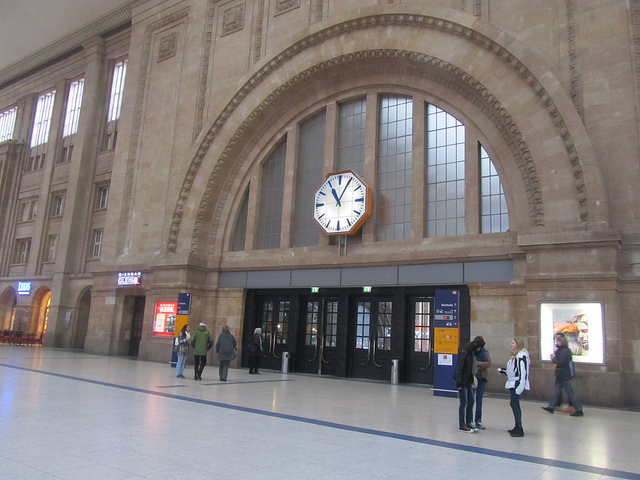}} & What is a famous example of the item in the middle of the picture? & clock & 2/5 & - \\\\\\\\
        \addlinespace\midrule
        \multirow{2}{*}{\includegraphics[height=2cm]{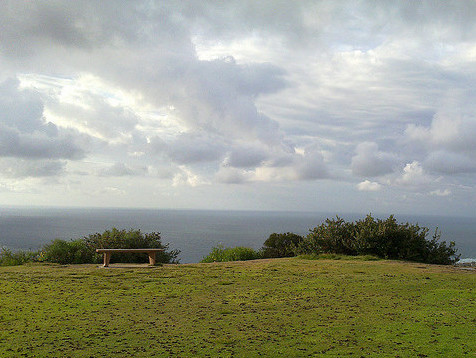}} & What time of day is it? & noon & 2/5 & You really cannot tell from the picture. \\\\\\\\
        \addlinespace\midrule
        \multirow{2}{*}{\includegraphics[height=1.5cm]{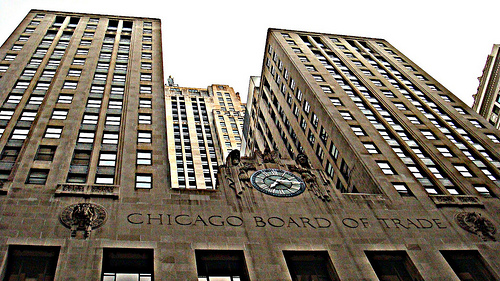}} & What color is the clock? & white & 2/5 & It is not white except for maybe small bits. \\\\\\
        \addlinespace\midrule
        \multirow{2}{*}{\includegraphics[height=1.8cm]{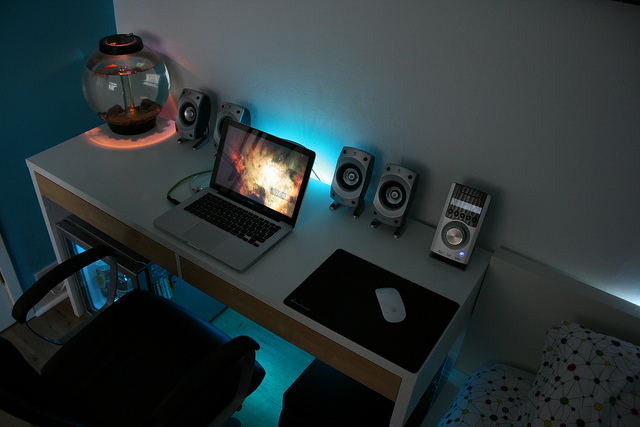}} & The picture on the laptop screen is called a what? & display & 2/5 & - \\\\\\\\
        \addlinespace\midrule
        \multirow{2}{*}{\includegraphics[height=2cm]{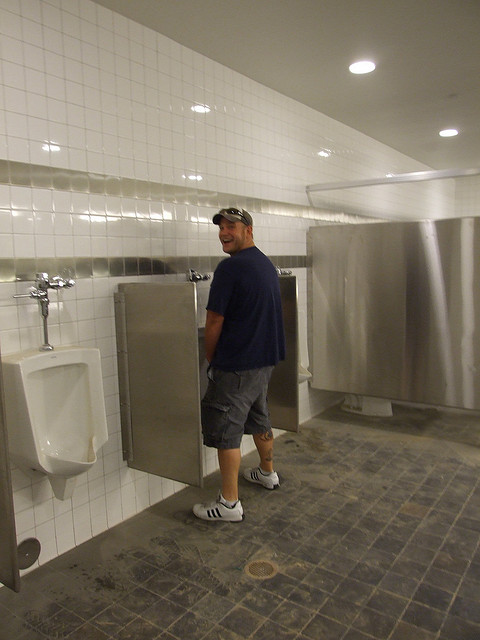}} & How many ceiling lights are there? & 2 & 3/5 & 2 lights are directly visible and a third is only seen as a reflection. \\\\\\
        \addlinespace\midrule
        \multirow{2}{*}{\includegraphics[height=2cm]{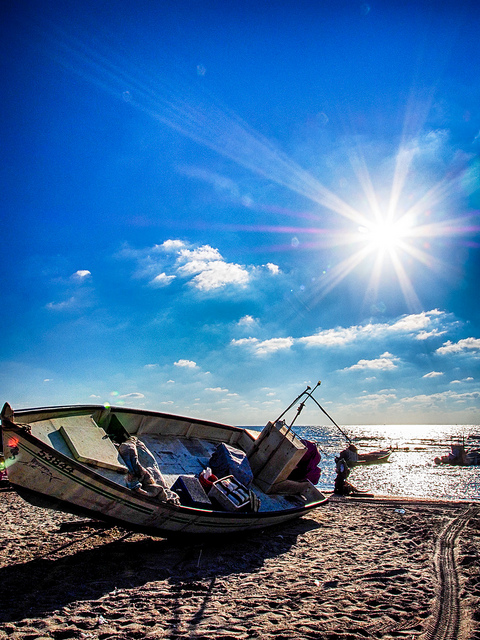}} & What is inside the boat? & luggage & 3/5 & Unanswerable from just the picture. \\\\\\\\
        \addlinespace\midrule
        \multirow{2}{*}{\includegraphics[height=2cm]{figures/COCO_val2014_000000571585.jpeg}} & What is this machine used for? & refrigerator & 3/5 & - \\\\\\\\\\
        \addlinespace\midrule
        \multirow{2}{*}{\includegraphics[height=1.8cm]{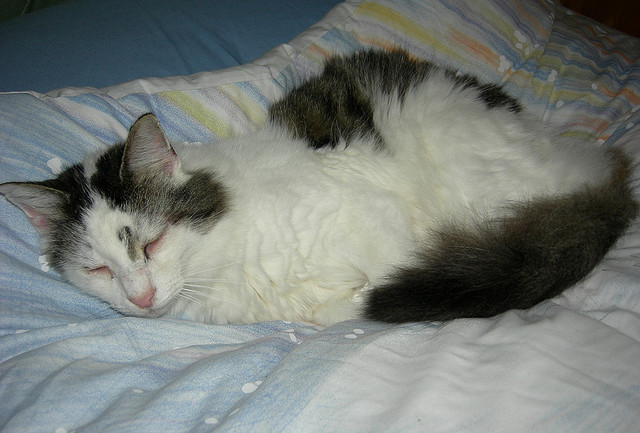}} & What do cats do this so much? & sleep & 3/5 & The question is confusing. \\\\\\\\
        \addlinespace\midrule
        \multirow{2}{*}{\includegraphics[height=1.8cm]{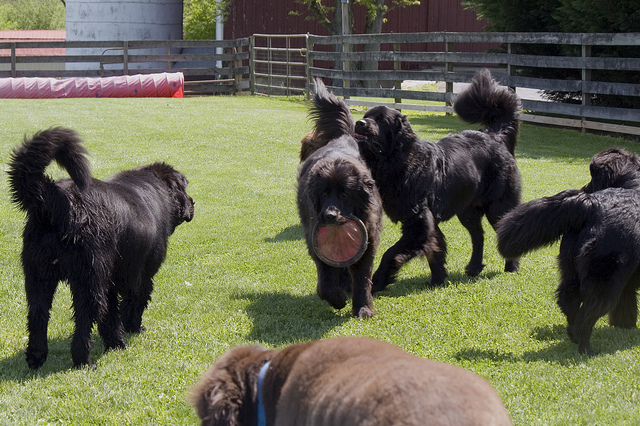}} & What animal are the animals in the picture known for not getting along with? & cats & 3/5 & - \\\\\\
        \bottomrule
    \end{tabular}
    \label{tab:annotator_disagreement}
\end{table*}

\subsection{Sets of Demonstrations for \Method{}}
\label{sec:demos}

Tab.~\ref{tab:demos_nonbinary} and Tab.~\ref{tab:demos_binary} contain the two sets of demonstrations used in \Method{}'s prompt.

\begin{table*}[ht]
\centering
\caption{Demonstrations with non-binary questions for \Method{}'s prompt.}
\begin{tabular*}{\linewidth}{@{\extracolsep{\fill}} l }
\toprule
\begin{lstlisting}[basicstyle=\fontsize{8}{10}\ttfamily, numbers=none]
Question: 'What is the color of the car?'
Reference answers: 'red', 'red', 'red', 'red', 'scarlet'
Candidate answer: 'pink'
Output: The candidate answer is incorrect because the car is 'red' and not 'pink'. So rating=1

Question: 'What is the animal on the left?'
Reference answers: 'elephant', 'giraffe', 'giraffe', 'giraffe', 'giraffe'
Candidate answer: 'giraffe'
Output: The candidate answer is correct because most of the reference answers (4 out of 5) indicate the animal on the left is a giraffe. So rating=3

Question: 'What's the weather like?'
Reference answers: 'bright', 'bright and sunny', 'clear', 'sunny', 'sunny', 'sunny'
Candidate answer: 'cloudy'
Output: The candidate answer is incorrect because the weather is 'bright' and 'sunny', not cloudy. So rating=1

Question: 'What are the people in the picture doing?'
Reference answers: 'sitting', 'sitting', 'sitting', 'sitting'
Candidate answer: 'they are resting'
Output: The candidate answer is ambiguous because, while it is common that people who are sitting are resting, it is not always the case. So rating=2

Question: 'What color are the base tiles?'
Reference answers: 'beige', 'beige', 'beige', 'brown', 'brown', 'tan', 'tan', 'tan', 'tan', 'ten'
Candidate answer: 'brown'
Output: The candidate answer is correct because the reference answers include 'brown' and other similar colors such as 'tan' or 'beige'. So rating=3

Question: 'How many people are in the picture?'
Reference answers: 'four', 'three', 'three', 'three', 'two', 'two'
Candidate answer: 'a few'
Output: The candidate answer is incomplete because 'a few' is less specific than the numerical reference answers. So rating=2

Question: 'What type of fruit is in the picture?'
Reference answers: 'apple'
Candidate answer: 'fruit'
Output: The candidate answer is incorrect because it does not specify the type of fruit. So rating=1

Question: 'What type of sculpture is this?'
Reference answers: 'Horse statue.'
Candidate answer: 'horse'
Output: The candidate answer is correct because 'horse' is equivalent to 'horse statue' in this context. So rating=3
\end{lstlisting} \\
\bottomrule
\end{tabular*}
\label{tab:demos_nonbinary}
\end{table*}

\begin{table*}[ht]
\centering
\caption{Demonstrations with binary questions for \Method{}'s prompt.}
\begin{tabular*}{\linewidth}{@{\extracolsep{\fill}} l }
\toprule
\begin{lstlisting}[basicstyle=\fontsize{8}{10}\ttfamily, numbers=none]
Question: 'Is the man wearing skis?'
Reference answers: 'yes', 'yes', 'yes', 'yes', 'yes', 'yes', 'yes', 'yes', 'yes', 'yes'
Candidate answer: 'yes'
Output: The candidate answer is correct because all the reference answers indicate the man is wearing skis. So rating=3

Question: 'Does the boy look happy?'
Reference answers: 'no', 'no', 'no', 'no', 'no', 'no', 'no', 'yes', 'yes', 'yes'
Candidate answer: 'smiling'
Output: The candidate answer 'smiling' is incorrect because it does not address the binary question, it should be either 'yes' or 'no'. So rating=1

Question: 'Is there a dog in the picture?'
Reference answers: 'no', 'no', 'no', 'no', 'no', 'yes', 'yes', 'yes', 'yes', 'yes'
Candidate answer: 'yes'
Output: The candidate answer is ambiguous because the reference answers are split between 'yes' (5) and 'no' (5). So rating=2

Question: 'Are these bears in their natural habitat?'
Reference answers: 'no', 'no', 'no', 'no', 'no', 'no', 'no', 'no', 'no', 'yes'
Candidate answer: 'yes'
Output: The candidate answer is incorrect because it contradicts the majority of the reference answers (9 out of 10), which indicate the bears are not in their natural habitat. So rating=1

Question: 'Is there a mountain in the picture?'
Reference answers: 'no', 'no', 'no', 'no', 'no', 'no', 'no', 'no', 'yes', 'yes'
Candidate answer: 'no'
Output: The candidate answer is correct because most of the reference answers (8 out of 10) indicate there is no mountain in the picture. So rating=3

Question: 'Is this a bathroom?'
Reference answers: 'no', 'yes', 'yes', 'yes', 'yes', 'yes', 'yes', 'yes', 'yes', 'yes'
Candidate answer: 'bathroom'
Output: The candidate answer 'bathroom' is incorrect because it does not address the binary question, it should be either 'yes' or 'no'. So rating=1

Question: 'Is this where boats are supposed to be?'
Reference answers: 'no', 'no', 'no', 'no', 'no', 'no', 'yes', 'yes', 'yes', 'yes'
Candidate answer: 'no'
Output: The candidate answer is ambiguous because the reference answers are split between 'yes' (4) and 'no' (6), indicating discrepancy about whether boats are supposed to be there. So rating=2

Question: 'Is the book on the table?'
Reference answers: 'no', 'no', 'no', 'yes', 'yes', 'yes', 'yes', 'yes', 'yes', 'yes'
Candidate answer: 'yes'
Output: The candidate answer 'yes' is correct because the majority of reference answers (7 out of 10) indicate the book is on the table. So rating=3
\end{lstlisting} \\
\bottomrule
\end{tabular*}
\label{tab:demos_binary}
\end{table*}

\subsection{Additional Correlation Results}
\label{sec:additional_results}

\begin{table*}[ht]
    \centering
    \caption{Kendall correlation ($\tau$) between VQA metrics and human judgment.}
    \resizebox{\textwidth}{!}{
    \begin{tabular}{lcccccccccc}
        \toprule
         & \multicolumn{3}{c}{BLIP-2} & \multicolumn{2}{c}{PromptCap} & \multicolumn{2}{c}{BLIP\textsubscript{VG}} & \multicolumn{2}{c}{BLIP\textsubscript{VQA}} & \multirow{2}{*}{Overall} \\
         & VQAv2 & VG-QA & OK-VQA & VQAv2 & OK-VQA & VQAv2 & OK-VQA & VG-QA & OK-VQA & \\
        \midrule
        \textbf{Baselines} \\
        VQA Acc. & 66.19 & 39.99 & 44.80 & 61.07 & 44.29 & 79.49 & 65.05 & 40.59 & 62.89 & 55.75 \\
        Soft VQA Acc. & 65.51 & 42.63 & 40.24 & 61.00 & 45.77 & 74.74 & 54.84 & 44.14 & 55.86 & 54.63 \\
        METEOR & 59.68 & 42.48 & 45.12 & 54.16 & 47.65 & 79.32 & 66.52 & 46.76 & 63.21 & 52.06 \\
        CIDEr & 60.01 & 41.65 & 44.74 & 54.14 & 42.28 & 76.40 & 63.20 & 41.85 & 61.06 & 55.08 \\
        BERTScore & 42.33 & 9.39 & 31.37 & 34.73 & 35.37 & 58.13 & 47.33 & 12.56 & 48.52 & 25.30 \\
        S-BERTScore & 51.89 & 33.35 & 39.40 & 41.48 & 39.75 & 64.24 & 55.25 & 37.88 & 53.68 & 46.18 \\
        \midrule
        \textbf{Ours} \\
        \Method{}~\textsubscript{FT5} & 68.34 & \textbf{58.16} & 57.25 & 61.73 & \textbf{56.54} & 69.08 & 62.82 & 52.77 & 60.46 & 62.14 \\
        \Method{}~\textsubscript{Vicuna} & 69.39 & 49.21 & 55.45 & 64.53 & 52.50 & 74.46 & \textbf{66.94} & 45.90 & \textbf{63.97} & 60.64 \\
        \Method{}~\textsubscript{GPT-3.5} & \textbf{71.17} & 57.68 & \textbf{58.07} & \textbf{69.23} & 54.79 & \textbf{80.46} & 65.57 & \textbf{56.21} & 62.75 & \textbf{65.31} \\
        \bottomrule
    \end{tabular}
    }
    \label{tab:kendall_results}
\end{table*}

Tab.~\ref{tab:kendall_results} contains Kendall correlations between VQA metrics and human judgments on every evaluated pair of VQA model and dataset. Some additional observations:

\textbf{Simple soft VQA Accuracy provides a decent improvement over traditional VQA Accuracy}, while being computationally less expensive than an LLM-based metric. Indeed, some failure modes of VQA Accuracy can be fixed by further standardizing (e.g. with character-level edit operations) the candidate and the reference answers before comparison. This suggests a tradeoff between computational cost and faithfulness to human judgment. However, Soft VQA Accuracy fails in cases where candidate and reference answers exhibit few-character differences in form yet substantial differences in meaning (e.g., digit-based answers in counting questions).

\textbf{There is a wider correlation gap between the baselines and \Method{} on VG-QA}. This is because VG-QA questions have a single reference answer, so metrics which use stricter comparisons between candidate and reference answers might miss correct candidates (false negatives) more frequently. Instead, using an LLM allows to recover some of these false negatives by reasoning about the semantic equivalence between reference and candidate answers.

\textbf{BERTScore does not perform as well as expected}. We believe this is because BERTScore computes a similarity score for each token in the candidate sentence with each token in the reference sentence, so it is meant to evaluate full sentences instead of short answers. S-BERTScore, based on global Sentence-BERT embeddings, has a higher correlation with human judgments compared to BERTScore, but it still underperforms other baselines, including n-gram based metrics such as METEOR or CIDEr.

\begin{table*}[ht]
    \centering
    \caption{Spearman ($\rho$) and Kendall ($\tau$) correlation between VQA metrics and human judgment, averaged across all VQA models and datasets.}
    \begin{tabular}{lcccccc}
        \toprule
         & \multicolumn{2}{c}{All} & \multicolumn{2}{c}{Binary} & \multicolumn{2}{c}{Other} \\
         & $\rho$ & $\tau$ & $\rho$ & $\tau$ & $\rho$ & $\tau$ \\
        \midrule
        \textbf{Baselines} \\
        VQA Acc. & 60.13 & 55.75 & \textbf{91.37} & \textbf{87.13} & 56.64 & 52.38 \\
        Soft VQA Acc. & 63.91 & 54.63 & 89.56 & 84.14 & 59.33 & 50.12 \\
        METEOR & 58.68 & 52.06 & 86.59 & 83.19 & 55.35 & 48.72 \\
        CIDEr & 63.88 & 55.08 & 89.09 & 81.04 & 60.62 & 51.99 \\
        BERTScore & 31.47 & 25.30 & 75.76 & 63.21 & 28.93 & 23.16 \\
        S-BERTScore & 56.61 & 46.18 & 83.62 & 73.43 & 52.53 & 42.43 \\
        \midrule
        \textbf{Ours} \\
        \Method{}~\textsubscript{FT5} & 64.99 & 62.14 & 78.32 & 75.10 & 63.24 & 60.51 \\
        \Method{}~\textsubscript{Vicuna} & 64.05 & 60.64 & 78.23 & 74.99 & 61.78 & 58.42 \\
        \Method{}~\textsubscript{GPT-3.5} & \textbf{68.91} & \textbf{65.31} & 88.40 & 85.18 & \textbf{65.72} & \textbf{62.17} \\
        \bottomrule
    \end{tabular}
    \label{tab:avg_correlation_results}
\end{table*}

\subsubsection{Correlation on Binary Questions}
\label{sec:binary_questions}

Tab.~\ref{tab:avg_correlation_results} shows the correlation breakdown between binary/yes-no questions ($12.57\%$) and other questions ($87.43\%$), averaged across all VQA models and datasets.
We notice a decreased performance of \Method{} on binary questions. Upon close inspection,
we find a large number of binary questions ($38.42\%$) contain both ``yes'' and ``no'' among the reference answers.
This contradiction results in a noisy/ambiguous context for the LLM, so the generated answer rating tends to be less reliable or consistent. Furthermore, when the candidate answer to a binary question is something other than ``yes'' or ``no'' (e.g. Q: ``Is the floor made of wood?'', A: ``tile''), the LLM might interpret the candidate as a plausible answer while humans label it as incorrect.
To help mitigate this issue, we appended ``\texttt{THIS IS VERY IMPORTANT: A binary question should only be answered with 'yes' or 'no', otherwise the candidate answer is incorrect.}'' to the prompt, but it only slightly improved performance with GPT-3.5.

\subsection{Comparison of Model Rankings Using VQA Accuracy vs. \Method{}}

\begin{figure}[ht]
    \centering
    \includegraphics[width=0.48\textwidth]{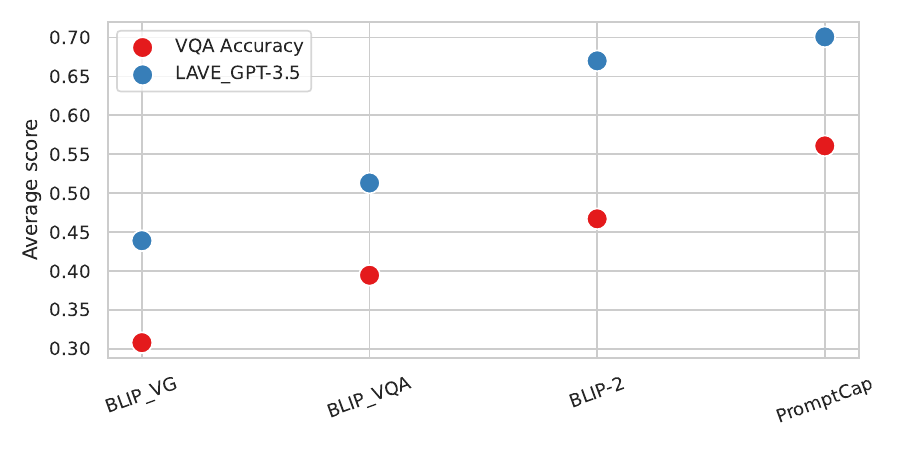}
    \caption{Average scores on OK-VQA ranked by model.}
    \label{fig:model_ranking}
\end{figure}

We compared the rankings of the considered models on the unseen OK-VQA benchmark using VQA Accuracy vs. \Method{}. We observe a consistent ranking across the two metrics (Figure~\ref{fig:model_ranking}).
Even though the rankings remain unchanged, the rate of change of scores is different for the two metrics. In particular, as per \Method{}, BLIP-2 and PromptCap are more similar to each other than as judged by VQA Accuracy. So these metrics are indeed highlighting different model behaviours.
Furthermore, although it makes sense to use VQA Accuracy during model development (due to its computational efficiency), we argue that \Method{} should be used for final model benchmarking because 1) it provides a more realistic measure of the state-of-the-art in VQA, 2) it helps understand where the real failures of VQA models lie and thus where future research should focus on, and 3) it provides rationales for its judgments (see Table~\ref{tab:qualitative_analysis}), thus making it more interpretable.

\subsection{Additional Qualitative Analysis: Is Visual Context Helpful?}
\label{sec:visual_context}

Since we are dealing with the task of \textit{visual} question answering, questions are inherently formulated in the context of an image. Hence, this visual context could potentially be leveraged for answer verification. A straightforward way of feeding visual context to an LLM is through an image description or caption (Sec.~\ref{sec:ablations}). However, relevant visual information is already encapsulated in the reference answers since human annotators observed the image in order to write them. This naturally leads to the question of how often visual context is really needed. To shed some light on this, we manually annotated 100 examples from four different model-dataset pairs --- (BLIP-2, VQAv2), (BLIP-2, VG-QA), (BLIP-2, OK-VQA), and (PromptCap, VQAv2) --- by asking ourselves if we feel confident in judging the model response solely by comparing it with reference answers, without consulting the image. We found that visual context is only needed in 12 out of 100 examples, while being helpful, albeit not strictly needed, in 11 examples. These are cases where annotators missed some valid answers to the question. Therefore, the image description generally does not offer additional information to verify the candidate answer beyond what is already contained in the reference answers.

\subsection{Limitations}
\label{sec:limitations}

An LLM-based VQA metric suffers from some of the same problems inherent to the underlying LLM. For instance, it has been observed that LLMs can hallucinate incorrect statements~\citep{maynez2020faithfulness,wei2022chain}. In our qualitative evaluation, we found this is occasionally the case for \Method{}, especially with ambiguous questions which contain multiple unique reference answers, sometimes even contradictory (e.g., ``yes'' and ``no''). However, in the context of automatic VQA evaluation, we believe the general flexibility and robustness of LLMs to different answer phrasings outweigh sporadic hallucinations.

Another limitation of \Method{} is its runtime. Using an LLM to produce rationales along with ratings makes evaluation slow due to the sequential nature of autoregressive generation. This issue can be partially mitigated with batched generation, although it requires high-end GPUs with a decent amount of VRAM. Another solution is to employ model compression techniques (quantization, pruning, distillation, etc.), which we leave for future work.
Nonetheless, when selecting a metric, we need to trade-off computational efficiency with accuracy. On one end of the spectrum, we have the existing VQA Accuracy, which is fast to compute but does not align closely with human judgment. On the opposite end, human evaluation serves as the gold standard but is prohibitively expensive and not scalable. We see \Method{} as a middle ground achieving a reasonable trade-off between computational cost and metric accuracy. Thus, we suggest using VQA Accuracy during model development and \Method{} for model benchmarking.

The best-performing variant of our metric, \Method{}~\textsubscript{GPT-3.5}, requires access to a paid API, which may preclude widespread adoption. Our motivation for using GPT-3.5 was to evaluate the effectiveness of our approach with one of the best LLMs currently available. However, we have shown that variants of our metric that use publicly available LLMs still outperform previous VQA metrics. We expect this issue to diminish as stronger instruction-tuned LLMs are publicly released. Nevertheless, using \Method{}~\textsubscript{GPT-3.5} to evaluate answer quality remains considerably cheaper than conducting human studies.

\end{document}